# Toward Measuring Structural Drift in LLM Communication Loops

Wael Hafez[1], Amir Nazeri[1], Chenan Wei[2]

[1] Semarx Research LLC, Alexandria, VA, USA

[2] University of Massachusetts, Amherst, MA, USA

Correspondence: w.hafez@semarx.com

## Abstract

Large language models increasingly run in stateful pipelines that assemble each prompt from retrieval, memory, tools, and other agents. Such pipelines drift: information that should shape the next response is dropped, compressed, or misrouted while every component still reports success. Existing diagnostics miss this because they evaluate isolated prompts, responses, or task scores, whereas what decouples is the relation between a prompt and the response it draws. Here we show that treating the prompt-to-response-to-next-prompt chain as the fundamental unit of analysis makes these relations measurable. We introduce *structural communication coherence*, quantified by two metrics: *communication closure*, which asks if what the pipeline returns at one turn matches what it faces next, and *normalized conditional action contribution*, which measures how much a sent message resolves the subsequent reply. Across 2,171 human–human, 58 human–LLM, and 8 LLM–LLM dialogues, these metrics reveal directional interaction structures; crucially, the measured contribution drops by 87–92% when a response is swapped for one from another turn, leaving surrounding prompts untouched. Because this approach requires no labels, healthy reference data, or predefined rules—only the raw prompts and responses—drift can be defined and measured directly from operational traffic, rather than inferred from eventual task failure. Establishing prospective detection performance is the next step.

## Introduction

Production LLM systems increasingly work as recurrent, multi-component interactions rather than isolated prompt-response calls. A prompt may combine conversation history, retrieval, persistent memory, tool outputs, workflow state, system instructions, human input and other agents. The response may then be parsed, routed, stored, executed or handed to another stage before the next call [19,23]. The behavior of the system is therefore produced by the interaction among stages, not by the model output alone.

This creates a drift problem that is different from component failure. Information that should shape what happens next can be omitted, compressed, replaced, misrouted or become less influential without any stage reporting an error. Recent work shows constraint loss under compaction, degradation in long-running memory and multi-agent interaction, and cases where context remains available but stops being useful [29-33,38-43]. Other methods observe drift through task scores, outputs, traces, rules or fitted reference behavior [16-18,24-28,34-37,44,45]. These methods are useful, but they look at different parts of the system.

Table 1 shows this difference in unit of analysis. SCC does not replace these methods. It adds a loop-level view.

| Approach | Primary unit observed | What it captures | Requires / cost | May miss |
|---|---|---|---|---|
| Semantic / hallucination evaluation | Output | Meaning, factuality, acceptability | Evaluator, prompts or task rules | Upstream decoupling when the output still looks plausible |
| Long-context / memory benchmarks | Task outcome | Retrieval, memory and reasoning degradation | Benchmark, labels or ground truth | Drift before the tested score moves, or outside |

| Approach | Primary unit observed | What it captures | Requires / cost | May miss |
|---|---|---|---|---|
| | | | | the benchmark task |
| Trace observability | Component / stage | Calls, context, retrieval, tools, routing | Instrumentation and diagnosis | Relations defined only between stages |
| Runtime contracts / constraints | Declared rule | Known constraint violations | Application-specific specifications | Undeclared or unexpected structural drift |
| Learned / reference monitors | Trajectory vs reference | Departure from fitted healthy behavior | Reference data and calibration | Drift outside the sampled reference regime |
| SCC (this work) | The prompt–response–next-prompt chain | Closure and directional contribution | The conversation's own prompts and responses; a declared representation for ε_A | Semantic correctness, task success and causal localization |

Table 1 | Current approaches observe different parts of LLM drift. SCC changes the unit of analysis to the prompt–response–next-prompt chain; it does not replace component-level, semantic or task-level methods.

The chain this paper analyzes runs from prompt to response to next prompt. When it decouples — when a response stops reflecting the prompt that drew it, or the next prompt stops carrying what the response returned — the next output becomes less predictable from the information that should have shaped it. Analyzing that requires the whole chain as the unit, not a single prompt, a single response or a single component. This follows the logic of communication theory: Shannon separates the technical structure of communication from meaning [1]; Weaver separates transmission, meaning and effects on behavior [2]; cybernetics places communication inside recurrent feedback [3,4]; and dialogue research treats communication as reciprocal adjustment across turns [5-7].

Two questions follow. Communication closure asks whether what the pipeline returns at one turn is what it faces at the next: whether the response, after any declared transformation, is what actually reaches the following prompt. Normalized conditional action contribution asks how much the message just sent resolves of what the other side says next, once that side's own previous message is taken into account — how much a prompt shapes the response it draws, and how much that response shapes the prompt that follows. A summary of one prompt or one response does not by itself define either relation, although a sufficiently joined trace contains the records from which both can be computed. Neither quantity says whether the answer is correct or the task succeeded.

The second quantity is not mathematically new. Its numerator is conditional mutual information. The normalized ratio is the conditional analogue of the uncertainty coefficient and, under a one-step dynamical interpretation, has the form of normalized transfer entropy [9-15]. The contribution here is the loop-level construction: closure across transitions, contribution within transitions, and both read in both directions of the loop.

## Results

### A recurrent communication loop has two directional records

Consider the alternating sequence $p_1$ $r_1$ $p_2$ $r_2$ $p_3$ $r_3$ …, where p_t is the prompt presented to the model at turn t and r_t is the returned response. Either side may itself be composite; for example, a human plus an application, retrieval layer, memory and tools can be treated as one participant if the other side receives one assembled message [19,23].

| side | faced message | action | next returned message |
|---|---|---|---|
| model side | p_t | r_t | p_t+1 |
| user side | r_t | p_t+1 | r_t+1 |

The same interaction therefore gives two records offset by half a turn. Every message is both one side's action and part of what the other side faces next. The recurrent communication loop - not a single prompt, response, model, retrieval step or tool call - is the primary object of analysis.

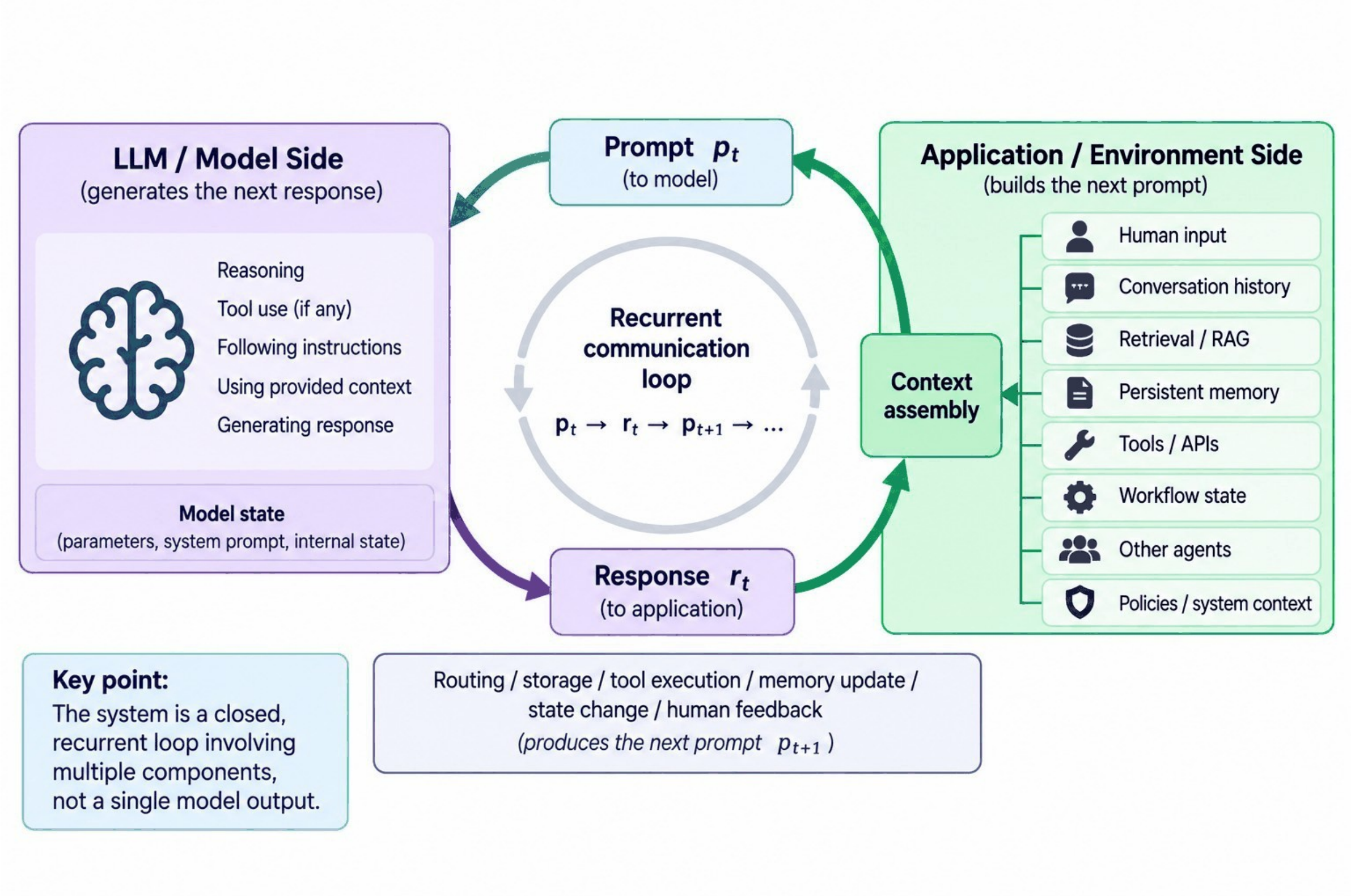

**Figure 1** The recurrent communication loop. The model side generates the next response from the prompt it faces. The application side builds the next prompt from history, retrieval, memory, tools, workflow state, human input, other agents and system context. Each message is both one side's action and the other side's next faced state. Schematic; no data.

## Closure is a property across transitions

Let S_t be the message or state faced by a stage and A_t the action it produces. Let R_t be the raw value returned on the next cycle, before any declared transformation, and let S_{t+1} be the state actually delivered to that stage at the next cycle. If a seam declares a transformation T, define Z_t = T(R_t). Closure requires

$$S_{t+1} = Z_t \, . \qquad (1)$$

Under an identity seam R_t = Z_t = S_{t+1} and the two coincide, which is the case in all three corpora here. Where T is lossy — compaction, redaction, filtering, routing — they differ, and every quantity below is defined against the delivered state S_{t+1}. We write S′ for that delivered state throughout. Over W transitions, finite-record continuity leaves only endpoint imbalance (Methods). We summarize the seam with

$$\Omega_T = H(Z) - H(S) \, . \qquad (2)$$

For an alphabet of size m, eventwise closure gives the finite-record bound

$$|\hat{\Omega}_T| \le h_2(\delta) + \delta \log(m-1), \quad \delta \le 1/W \, . \qquad (3)$$

The general rate is O(log W/W). A clean O(1/W) rate needs fixed-support regularity. Equality of the time-indexed population marginals, and therefore Ω_T=0 at population level, also needs stationarity or an equivalent repeated-realization assumption. $\Omega_T=0$ does not prove closure. Finer seam readings - eventwise mismatch and Jensen-Shannon divergence - have smaller blindness classes. Designed transformations such as compaction or redaction must be included in T [20,29-31].

## Conditional contribution is a property within transitions

Closure asks whether the loop is connected across transitions. Contribution asks whether one side still helps determine what follows within a transition. For a record $(S,A,S')$,

$$H(S') = I(S;S') + I(A;S'|S) + H(S'|S,A) . \quad (4)$$

We define the normalized conditional action contribution as

$$\varepsilon_A = I(A;S'|S) \,/\, H(S'|S) = 1 - H(S'|S,A)/H(S'|S), \quad H(S'|S)>0 . \quad (5)$$

$\varepsilon_A$, the normalized conditional action contribution, is the share of residual uncertainty in the delivered next message, after the prior state is known, that is resolved by the action. For the population quantity, $0\leq\varepsilon_A\leq1$. If $H(S'|S)=0$, $\varepsilon_A$ is undefined, not zero. Contribution is defined against the delivered state; where the seam is lossy, contribution to the raw return R is a distinct quantity and is not reported here. The numerator has the form of transfer entropy under a first-order dynamical interpretation [9-15]. The quantity is observational and is not read causally. Each side has its own $\varepsilon_A$, so directional differences can be reported without collapsing the two sides into one score.

## Empirical behavior across three communication-loop types

We computed the quantities on three corpora: C1, 58 human-LLM conversations with 25-1,402 paired turns; C2, eight 25-turn LLM-LLM runs; and C3, 2,171 eligible human-human PhotoBook games [8]. These corpora illustrate the loop-level quantities; they do not establish the definitions or bounds.

### The closure bound is an identity and indexing check in these logs

In these corpora the seam is the identity map: the same logged message is both the value returned at one transition and the state faced at the next. Closure therefore holds by indexing. The finite-record bound is an identity and indexing check, not a test of fault detection. Testing an omitted, changed, compacted, redacted or misrouted message would require two separate seam taps - what was emitted upstream and what was received downstream - which these records do not provide.

As expected, no $\Omega_T$ value exceeds the bound at $W\geq2$. The observed maxima use only part of the allowed band and shrink with W. In C1 the prompt-side ratio falls from 0.531 at $W=2$ to 0.370 at $W=50$, while the response side falls from 0.403 to 0.219. C3 falls from 0.422 to 0.214; the shorter C2 records remain lower.

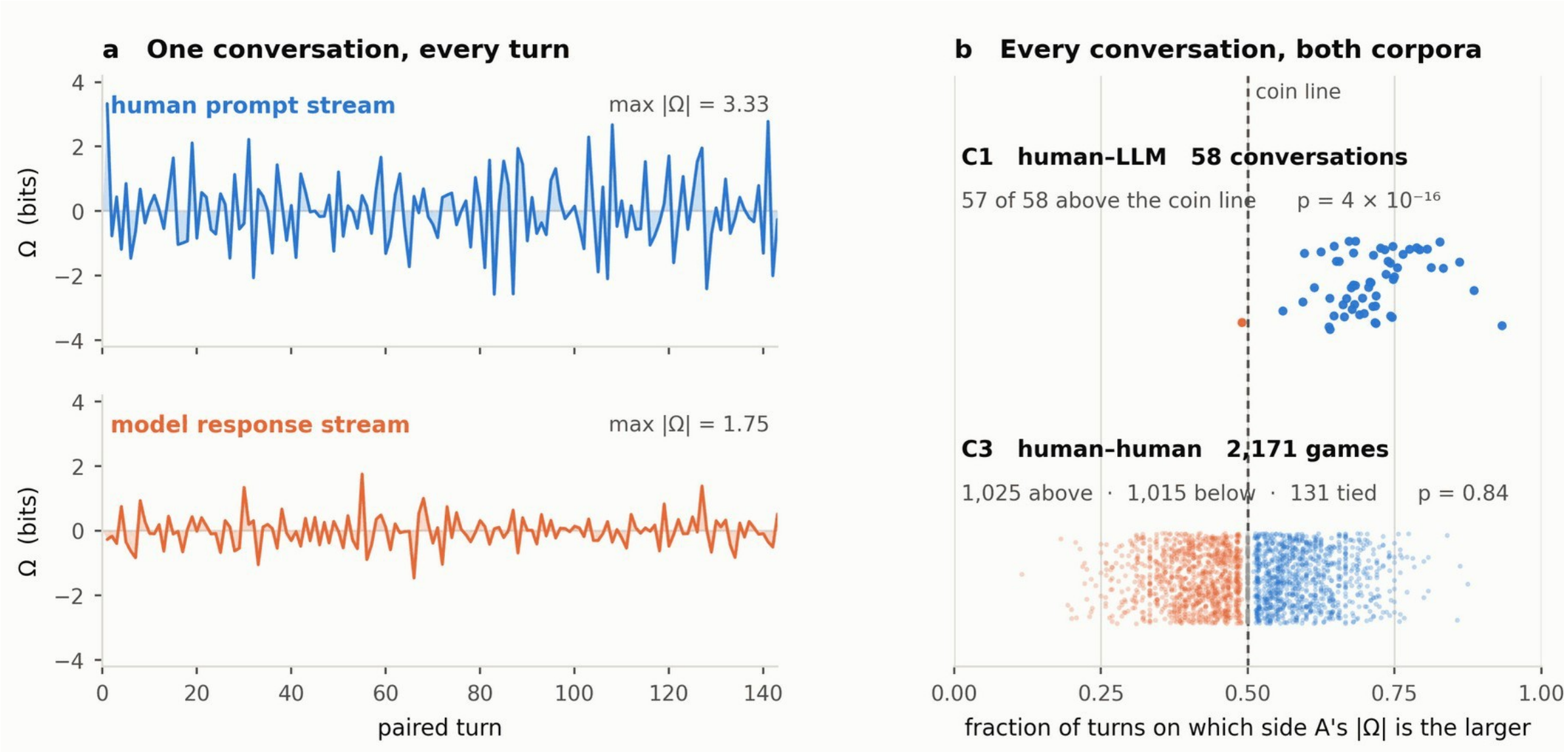

**Figure 2** Directional structure of endpoint closure fluctuations. a, $\Omega$ at W=1 for one C1 conversation. b, for each record, the fraction of turns on which side A has the larger $|\Omega|$. C1 is strongly one-sided and C3 is balanced. These panels describe endpoint token-distribution variability in correctly indexed identity-seam logs; they are not tests of closure detection.

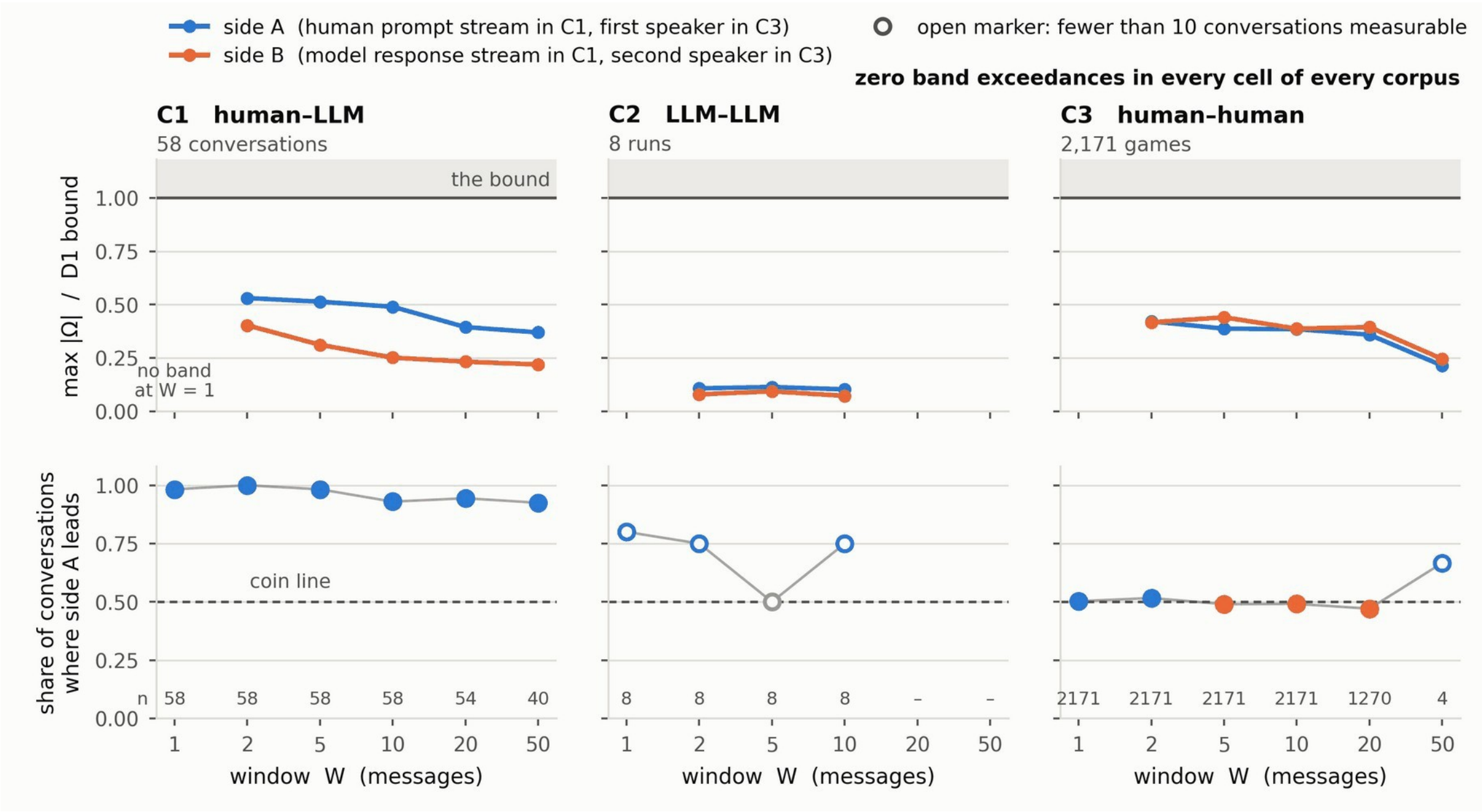

**Figure 3** Band occupancy and directional share across window lengths. Top row: largest observed $|\Omega_T|$ as a fraction of the bound. Bottom row: share of records in which side A has the larger excursion. No band is defined at W=1; open markers indicate fewer than ten measurable records.

## Closure fluctuations are directional

In C1, one side has the larger excursion in 57 of 58 conversations, and the pattern holds at every tested window length. C3 is balanced (1,025 versus 1,015, with 131 ties). C2 is too small to resolve a directional pattern. The narrow conclusion is that the corpora have different directional profiles in endpoint token distributions. We do not show that these profiles track closure faults, because no closure fault is present in these logs.

## Conditional contribution is measurable in a subset of exchanges

$\varepsilon_A$ needs a declared symbol representation and a reference distribution. Under the conditional reference, 25 of 58 model-side and 36 of 58 user-side C1 readings clear, with both sides clearing in 24 conversations. Pooled C2 clears on both sides at $k=2$ and at $k=3$ (four readings, $N=184$ each). In C3, 620 of 2,171 second-speaker and 784 first-speaker readings clear, with both sides clearing in 143 games. Records that do not clear are inconclusive, not low; short records have less power.

Directional comparisons use only records where both sides clear. C3 is balanced: 74 of 143 games are negative (51.7%, $p=0.74$). In C1, $\hat{\varepsilon}_A(\text{model})-\hat{\varepsilon}_A(\text{human})$ is negative in 20 of 24 conversations, with median −0.055 and bootstrap 95% interval [−0.084, −0.029]. This comparison was exploratory. Of the 24 records, 22 come from one platform (18 of 22 negative, median −0.051) and two from another; 13 of 14 records above 300 turns are negative against 6 of 9 between 100 and 300. All 58 C1 conversations come from one author across three platforms, so this is a descriptive result on a selected subset, not evidence of a general human-versus-model pattern.

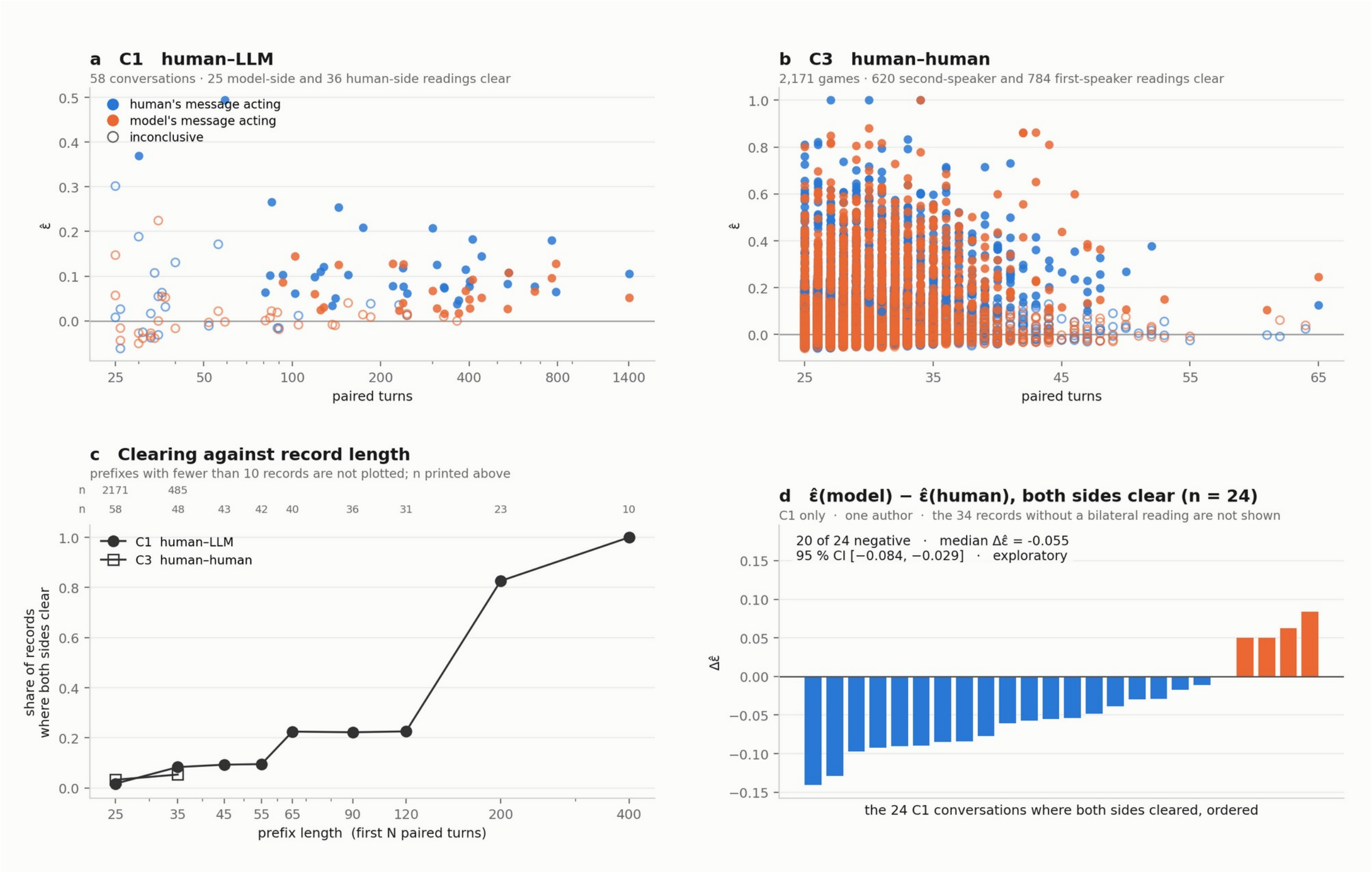

**Figure 4 |** Conditional contribution across corpora. a, C1: $\hat{\varepsilon}_A$ for each of 58 human–LLM conversations, both sides, against record length. Filled markers clear the conditional reference; open markers are inconclusive, which is a reading, not a missing value. b, C3: the same for 2,171 PhotoBook games. c, Share of records in which both sides clear, by record length; bins with fewer than ten records are not plotted and denominators are printed above. d, $\hat{\varepsilon}_A(\text{model}) - \hat{\varepsilon}_A(\text{human})$ for the 24 C1 conversations in which both sides clear, ordered; 20 are negative, median −0.055. This comparison is exploratory and all 58 C1 conversations come from one author. Blue: the human's message as the action (first speaker in C3). Orange: the model's message (second speaker in C3).

## Substituting the paired action removes most measured contribution

The controls replace each action with an action from another turn while keeping S and S′ fixed. This leaves $I(S;S')$ and $H(S'|S)$ unchanged, so movement in $\hat{\varepsilon}_A$ comes from the conditional contribution term.

The controls separate adjacency from action–state compatibility. Control C keeps the same S symbol and a local block of the record but breaks adjacency. Control A matches message length. Control B also forces low semantic similarity. Across ten C1 and C2 cells, median $\hat{\varepsilon}_A$ falls by 92.0% under Control C, 86.8% under Control A and 90.0% under Control B. Control C removes the most contribution while drawing substitutes that are, on average, the most similar to the real action — markedly so in the C2 cells. The reductions are therefore consistent with sensitivity to the observed action–transition pairing in the declared representation, beyond message length and coarse state symbol alone. They do not isolate immediate adjacency as the unique source: the state symbol is a coarse cluster, two observations sharing it may differ in finer topic or dialogue act, and the 92.0% and 90.0% medians are close. The controls remain observational and do not rule out dependence beyond the local block.

Because $\hat{\varepsilon}_A$ is built from Miller-Madow-corrected entropy estimates, the finite-sample estimate can fall outside [0,1] even though the population quantity cannot. Values are therefore left unclipped and are treated as estimator artifacts when they leave the population range.

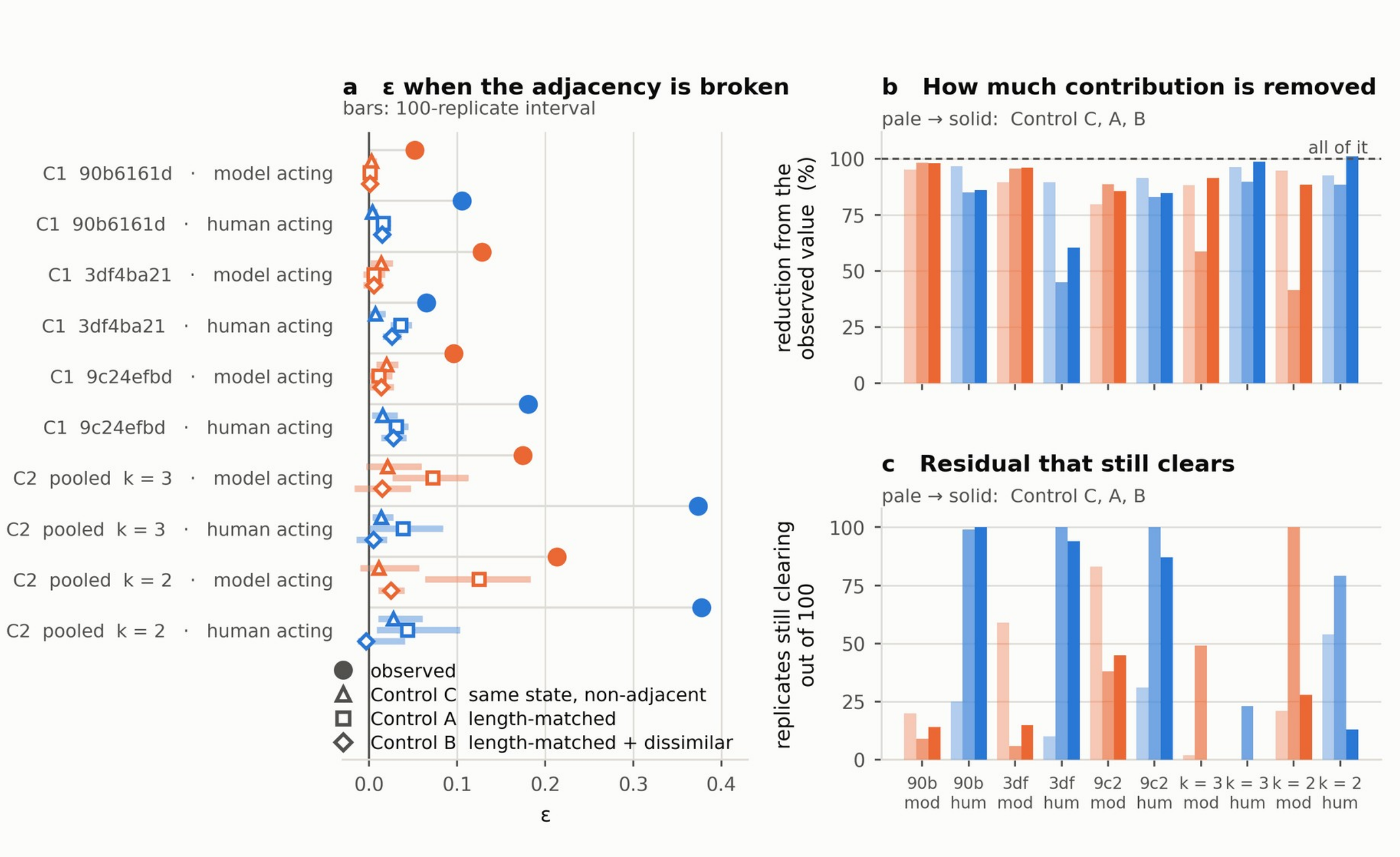

**Figure 5** Substituting the paired action removes most measured contribution. Each control replaces every action with one from another turn of the same record, keeping S and S′ fixed, so I(S;S′) and H(S′|S) are unchanged. a, Observed $\hat{\varepsilon}_A$ and the three controls across ten cells, with the 100-replicate interval. b, Reduction from the observed value: median 92.0% under Control C, 86.8% under Control A and 90.0% under Control B. c, Share of replicates that still clear the reference. Control C substitutes are on average the most similar to the real action (median cosine 0.53 in the C2 cells, against 0.48 for Control A and 0.37 for Control B) yet remove the most contribution. The reductions are consistent with sensitivity to the observed action–transition pairing in the declared representation; they do not isolate immediate adjacency as the unique source.

## Interaction-structural drift

Closure and conditional contribution use different references. Closure is compared with a derived structural constraint. Contribution has no universal healthy value and is read relative to the loop's established regime. We define interaction-structural drift as a change in closure across transitions, conditional contribution within transitions, or their directional structure across the two sides.

This is different from semantic drift or task failure. A pipeline can remain structurally coherent while doing the wrong task, and a structurally changed loop may still produce an acceptable answer. Loop-level readings are therefore candidate structural summaries that can motivate deeper inspection of retrieval, memory, context assembly, tool use, handoffs or workflow execution using existing traces [18,24,28]. The permutation reference asks whether conditional contribution is measurable in a record; it is not itself a drift baseline, which would require comparison against a declared regime across time.

## Discussion

Interaction-structural drift is a property of the prompt–response chain, not of its parts. Closure is defined across a turn, between what the pipeline returns and what it next faces. Contribution is defined between one message and what the other side says next. A stage-local summary does not by itself define either relation, though a trace that aligns the stages contains the records needed to compute it. This is a statement about the unit of analysis, not a claim that SCC outperforms existing monitoring methods.

This unit-of-analysis difference explains how SCC relates to prior work. Long-context and memory benchmarks show drift through retrieval, reasoning, memory use or task performance [38-43]. Semantic and hallucination evaluators show it through output quality [44]. Observability shows component traces [16-18,24-28], while contracts and learned monitors show rule violations or departure from fitted regimes [34-37]. SCC adds a loop-level structural view. The closure reference is analytically derived and does not need a healthy reference corpus.

$\Omega_T$ and $\varepsilon_A$ are examples of what becomes measurable once the recurrent loop is chosen as the unit of analysis. They do not solve production drift detection. $\Omega_T$ is coarse: equal entropy does not imply equal distributions or equal events. $\varepsilon_A$ depends on the chosen representation and short records have limited power. Neither quantity establishes semantic correctness, task success or causal influence.

The main claim is methodological: drift can be analyzed as a change in how much each prompt shapes the response it draws, and how much each response shapes the prompt that follows. This paper provides that unit of analysis and two candidate summaries. It does not provide a validated prospective detector. No labeled change point, predeclared alert threshold, sensitivity, specificity, false-alarm rate or detection delay is evaluated here. Because closure and conditional contribution are defined on interaction structure rather than on task success, they provide a natural basis for prospective runtime monitoring; a next step is to test whether changes in them precede or complement semantic, task-level and component-level indicators under controlled retrieval, memory, compaction and handoff failures. The construction is also per stage rather than per pair: in a pipeline with more than two stages, each stage has its own directional record, contribution reading and closure seam, so a departure can in principle be assigned to a stage rather than to the pipeline as a whole. We state that generalization without testing it here.

## Methods

All bounds, identities and null classes used in the main text are derived here. Citations are used only where a construction has an established name in the literature.

### Setting and notation

All variables take values in finite alphabets and all entropies are in bits. $H(\cdot)$ is Shannon entropy [1], $I(\cdot;\cdot)$ mutual information, $I(\cdot;\cdot|\cdot)$ conditional mutual information, $h_2$ the binary entropy function and JSD the Jensen-Shannon divergence. A finite record contains W transitions. Participant policies and representations are treated as fixed within that record. For a stage, S is the message/state faced and A the action/message produced. R is the raw value returned on the next cycle and S′ the state delivered at the next cycle, so that T is the declared seam transformation, Z=T(R) and closure requires S′=Z. For an identity seam R=Z=S′, which is the case in all corpora reported here. All contribution quantities are defined against the delivered state S′. The alphabet size in the closure bound is m. In the two-party LLM loop, $p_t$ is the prompt at turn t and $r_t$ the response.

## Closure

Let the record contain states $S_0,\dots,S_W$ with transitions $t=0,\dots,W-1$ and counts $C_{ij}=\#\{t:S_t=i,S_{t+1}=j\}$. Define $out(i)=\Sigma_j C_{ij}$ and $in(i)=\Sigma_j C_{ji}$.

$$out(i) - in(i) = 1\{S_0=i\} - 1\{S_W=i\}. \quad (6)$$

Every interior visit contributes once to both in- and out-counts; only the endpoints remain. This identity is exact and does not assume stationarity. Equality of ensemble marginals follows only under stationarity or repeated realizations.

Under closure, $S_{t+1}=Z_t$. The empirical faced and returned histograms are the same trajectory viewed one step apart, so their total variation is at most $\delta\leq 1/W$. By the sharp Fannes-Audenaert continuity bound for entropy, obtained in the classical setting by Zhang [46] and as Theorem 1 of Audenaert [47],

$$|\hat{\Omega}_T| \leq h_2(\delta) + \delta \log(m-1), \quad \delta\leq 1/W . \quad (3)$$

Therefore $|\hat{\Omega}_T|=O(\log W/W)$ in general. $O(1/W)$ follows when active probabilities stay bounded away from zero. Under stationarity or repeated realizations, the endpoint term vanishes in the population and $H(Z)=H(S)$, so $\Omega_T=0$. The converse does not hold: $\Omega_T=0$ does not establish closure. Closure alone does not define a universal finite-sample distribution for $\hat{\Omega}_T$; under mixing stationarity, a delta-method or moving-block bootstrap can be used.

## Seam instruments and transformations

For deterministic T, the eventwise mismatch rate is $m_T=\Pr[S'_t\neq Z_t]=\Pr[S'_t\neq T(R_t)]$, the distributional discrepancy is $JSD(p_Z,p_S)$, and the informational balance is $\Omega_T=H(Z)-H(S)$. Their nulls satisfy $m_T=0 \Rightarrow JSD=0 \Rightarrow \Omega_T=0$, with no converse. $\Omega_T$ is blind to entropy-preserving substitutions; JSD is blind to distribution-preserving substitutions.

For stochastic seams with designed kernel $T(y|x)$ and realized kernel $Q(y|x)$, a conditional divergence such as $d_T=\Sigma_x\, p(x)\, JSD(Q(\cdot|x),T(\cdot|x))$ is suitable. All seam instruments are written in the transformed coordinate $Z=T(R)$. If T is many-to-one, the intended entropy loss is part of the seam and is not a closure failure.

## Conditional action contribution

$$H(S') = I(S;S') + I(A;S'|S) + H(S'|S,A). \quad (4)$$

This follows by expanding $H(S')=I(S;S')+H(S'|S)$ and then $H(S'|S)=I(A;S'|S)+H(S'|S,A)$. For $H(S'|S)>0$,

$$\varepsilon_A = I(A;S'|S)/H(S'|S) = 1 - H(S'|S,A)/H(S'|S). \quad (5)$$

For the population quantity, $0\leq\varepsilon_A\leq 1$. The estimator uses Miller-Madow-corrected entropies before the ratio is formed, so $\hat{\varepsilon}_A$ can fall outside [0,1] on sparse tables. Estimates are not clipped; values outside the range are labeled estimator artifacts. If $H(S'|S)=0$, $\varepsilon_A$ is undefined. Near that boundary the ratio is unstable, which motivates the room screen. Also $I(A;S'|S)\leq H(A|S)$, so the numerator is zero if the action is fully determined by S.

The numerator has the first-order transfer-entropy form under the matching temporal interpretation [9]. The ratio is the conditional analogue of the uncertainty coefficient [10], with normalized-transfer-entropy precedents [11,12]. No causal interpretation is made; causal claims would need interventions or identification assumptions [13,14].

## Representation

Closure and contribution use different representations. For closure, each message is tokenized with the Llama-2 tokenizer (NousResearch/Llama-2-7b-hf; vocabulary K=32,000) and represented by a normalized token-frequency vector $q_i$. Over W messages,

$$\bar{q}_Z(t)=(1/W)\Sigma_{i=t-W+1}^{t}q_i, \quad (7)$$

$$\bar{q}_S(t)=(1/W)\Sigma_{i=t-W}^{t-1}q_i, \quad (8)$$

$$\hat{\Omega}_T(t)=H(\bar{q}_Z(t))-H(\bar{q}_S(t)). \quad (9)$$

The implemented quantity is the entropy of an equal-message-weighted token mixture, not the entropy of a distribution over message states; the alphabet size m in the bound is therefore K. Equal message weighting is required for $\delta \leq 1/W$. Empty messages are dropped. Fewer than ten $\hat{\Omega}_T$ values at a given W is unmeasurable at that W. No band is reported at W=1. The window grid is W=1,2,5,10,20,50 with step 1.

For contribution, messages are embedded with sentence-transformers/all-MiniLM-L6-v2, mean-pooled over the attention mask, L2-normalized, max_length=256, and assigned to symbols by k-means fitted within each declared record, with the two streams clustered separately (random_state=0, n_init=10). k=2 for 25-149 turns, k=3 for 150-399 and k=4 for 400 or more. The eligibility floor is 25 turns. One (S,A,S′) triple is formed per turn per side. Equal-frequency character-length bins at the same k provide a robustness check.

## Estimation, reference and screens

Closure and contribution use different estimators, because the closure band is derived for the empirical distributions themselves. Closure entropies are therefore plug-in, uncorrected, so that the reported quantity and the bound of equation (3) are the same object. Contribution entropies use the plug-in estimator with the Miller-Madow correction $\hat{H}+(\hat{m}-1)/(2N \ln 2)$, where $\hat{m}$ is the number of occupied cells and N the sample count, applied to each entropy before the ratio is formed. The primary reference preserves the conditional null: A is permuted only among transitions with the same S symbol, using strata of at least five transitions, and retained strata must cover at least 70% of transitions. In C1, median coverage is 1.00 and no record falls below the floor. In the pooled C2 analysis, permutation is additionally confined within each run, so actions never move between runs; within-run coverage is 0.86–0.96 and the pooled and run-preserving references agree on every cell. If strata are too sparse, A is permuted among the ten nearest neighbours of S in the pre-clustering embedding, following Runge [48]. A circular shift by at least ten transitions is used only as a serial-structure diagnostic. Each design uses 2,000 permutations, seeded from the md5 of record identifier, side and design. A value above the 95th percentile of its own reference is a clear reading; otherwise it is inconclusive.

Screens are applied before any value is read. A stream using fewer than k symbols fails symbol occupancy. Fewer than about three samples per occupied cell fails cell occupancy. $H(S'|S) \leq 0.05$ bits fails the room screen. $\hat{\varepsilon}_A \geq 0.99$ fails the saturation screen and is flagged, not treated as a strong reading. Fewer than ten $\hat{\Omega}_T$ values at a given W fails the minimum-readings screen. A failed screen is unmeasurable, never low. $\Delta\hat{\varepsilon}_A$ is reported only when both sides clear.

False-positive rate and power were calibrated by simulation across (N,k) combinations from (23,2) to (800,4), covering the full range of effective transition counts admitted by the 25-turn floor. Records were generated with S a Markov chain of declared self-transition probability and stationary skew, A drawn from p(A|S) at two dependence strengths, and S′ drawn from p(S′|S) under the null or from p(S′|S,A) under the alternative, spanning stationary and mid-record regime changes and balanced and skewed occupancy. The alternative applies a declared coefficient of 0.8 to the action term in the next-state logits, which produces a mean $\hat{\varepsilon}_A$ of 0.10–0.26 at short record lengths against 0.07 in the corresponding null cells. Long records use 200 simulations and 200 permutations per cell and short records 300; per-cell counts and Wilson intervals are reported in the released calibration table. The primary within-state design holds or beats nominal level throughout. From N=40 upward its mean empirical type-I error is 0.045 (range 0.015–0.075), stable across action–state dependence at 0.0452 against 0.0453. At the shortest admitted lengths it is conservative rather than inflated: mean 0.028 across 23–39 transitions, 0.021 at N=23 with k=2 and 0.037 at N=23 with k=3, with a worst-cell Wilson upper bound of 0.097. Clearing counts on short records are therefore understated rather than overstated, and no record is excluded on calibration grounds. The local-permutation fallback behaves similarly (0.042–0.047 at short lengths). Global permutation is mildly anticonservative throughout (mean 0.054; worst 0.085). The circular surrogate degrades sharply on short records, from 0.064 at N=39 to 0.22 at N=23, which is why it is reported only as a serial-structure diagnostic and never as the reference. Power rises with record length at fixed effect size, from 0.14–0.24 across 23–39 transitions to 0.24 at N=40, 0.44 at N=100, 0.84 at N=250 and 1.00 at N=800; the rising frequency of clear readings with record length is therefore a property of the estimator.

## Corpora

C1 - human-LLM. Fifty-eight conversations between one author and deployed assistants across three platforms, with 25-1,402 paired turns. They were collected during ordinary work, not under an experimental protocol. Inclusion uses only the 25-turn floor. Because all records share one human participant, they are not independent experimental units; results are descriptive.

C2 — LLM–LLM. Eight 25-turn exchanges between an asking model and an answering model under one task and protocol, differing only by seed. The first turn of each run is excluded, as in the closure definition, leaving 24 turns and therefore 23 adjacent triples per run, or 184 per side across the eight runs. Because no run reaches the per-conversation alphabet floor, messages are embedded together and k-means is fitted once over the pooled prompt set and pooled response set. Triples are formed within each run and then pooled. k=3 is primary and k=2 is secondary.

C3 - human-human. The PhotoBook corpus [8]. Selection messages are removed, consecutive messages from the same speaker are collapsed, and non-overlapping first-speaker-to-second-speaker turns are formed. Six games with out-of-range round scores and 325 games below 25 turns are excluded, leaving 2,171 of 2,502 games with 25-65 turns.

## Pairing controls

All controls replace each action with one from another turn of the same record, keep S and S′ fixed, exclude adjacent turns ($|j-t|\geq 2$) and never cross a record boundary. $I(S;S')$ and $H(S'|S)$ are therefore unchanged. The controls cover the three longest C1 conversations at k=4 and pooled C2 at k=3 and k=2, with 100 replicates per cell. C3 is not used for these structural controls.

Control C selects a substitute from turns with the same S symbol and within ±15% of record length, preserving state compatibility and local phase while breaking adjacency. Control A chooses the nearest message length, with replicate draws within ±5%. Control B chooses among length-matched candidates below cosine similarity 0.50, widening the length band through ±10%, 20%, 50%, 100% and then unrestricted until a pool exists. Ties are broken by md5 seed. Estimator settings, screens and references are unchanged between original and controlled data.

## Reproducibility

All parameters were declared before computation and were not tuned per corpus. The one amendment lowered the eligibility floor from 60 to 25 turns on 14 September 2026. The 60-turn floor carried two requirements at once: a $k^3$ table needs enough samples per cell, and a sliding window of 40 turns needs a longer record. Only the first binds when $\varepsilon_A$ is read once per record rather than in a window, and at k=2 a 25-turn record gives 24 triples over eight cells, exactly three per cell, which is the cell-occupancy screen. The change was made on that calibration basis. The 18 conversations admitted by the lower floor contribute none of the 24 records in which both sides clear, so the directional comparison is identical under either floor. The directional comparison between sides was exploratory and is labeled as such. No multiplicity correction is applied. Replication would require an independent multi-user corpus. Permutation seeds are deterministic. Software versions, tokenizer, embedder, hardware and runtime configuration are recorded with the release package.

## Data and code availability

C3 is the public PhotoBook corpus [8]. C1 consists of one author's private conversations with deployed assistants, collected in ordinary use; that author is the only human participant and no third-party personal data is involved, and the conversation text is not released. Derived data, estimator implementations and analysis code are available from the corresponding author on reasonable request.